\documentclass[10pt,twocolumn,letterpaper]{article}

\usepackage{3dv}
\usepackage{times}
\usepackage{epsfig}
\usepackage{graphicx}
\usepackage{amsmath}
\usepackage{amssymb}
\usepackage{booktabs}
\usepackage{float}

\threedvfinalcopy 

\def\threedvPaperID{141} 

\ifthreedvfinal\fi
\begin{document}

\title{NodeSLAM: Neural Object Descriptors for Multi-View Shape Reconstruction}

\author{Edgar Sucar \hspace{1em} Kentaro Wada \hspace{1em} Andrew Davison\\
Dyson Robotics Laboratory, Imperial College London\\
{\tt\small \{e.sucar18, k.wada18, a.davison\}@imperial.ac.uk}
}

\maketitle

\begin{abstract}
   The choice of scene representation is crucial in both the shape inference algorithms it requires and the smart applications it enables. We present efficient and optimisable multi-class learned object descriptors together with a novel probabilistic and differential rendering engine, for principled full object shape inference from one or more RGB-D images. Our framework allows for accurate and robust 3D object reconstruction which enables multiple applications including robot grasping and placing, augmented reality, and the first object-level SLAM system capable of optimising object poses and shapes jointly with camera trajectory.
\end{abstract}

\section{Introduction}

To enable advanced AI applications, computer vision algorithms must build useful persistent 3D representations of the scenes they observe from one or more views, especially of the objects available for interaction. Ideal object representations are: (i) efficient, for principled and fast optimisation, (ii) robust to noisy measurements and variable uncertainties, and (iii) incremental, with the ability to grow and improve with new measurements. In this paper, we study the use of generative class-level object models and argue that they provide the right representation for principled and practical shape inference as shown in the experiments. We use a novel probabilistic rendering measurement function which enables efficient and robust optimisation of object shape with respect to depth images, and build from this an object-level SLAM system capable of incrementally mapping multi-object scenes.

\begin{figure}[htbp]
\begin{center}
\includegraphics[width=0.9\linewidth]{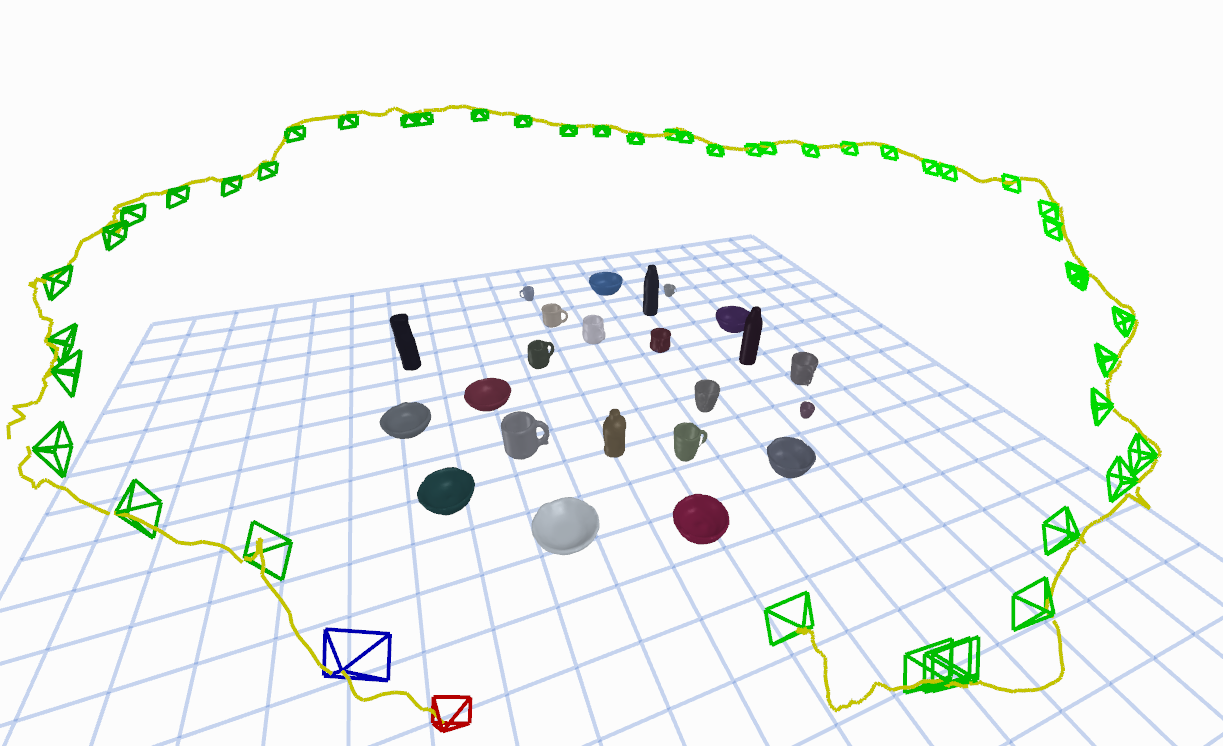}

\vspace{2mm}

\includegraphics[width=0.9\linewidth]{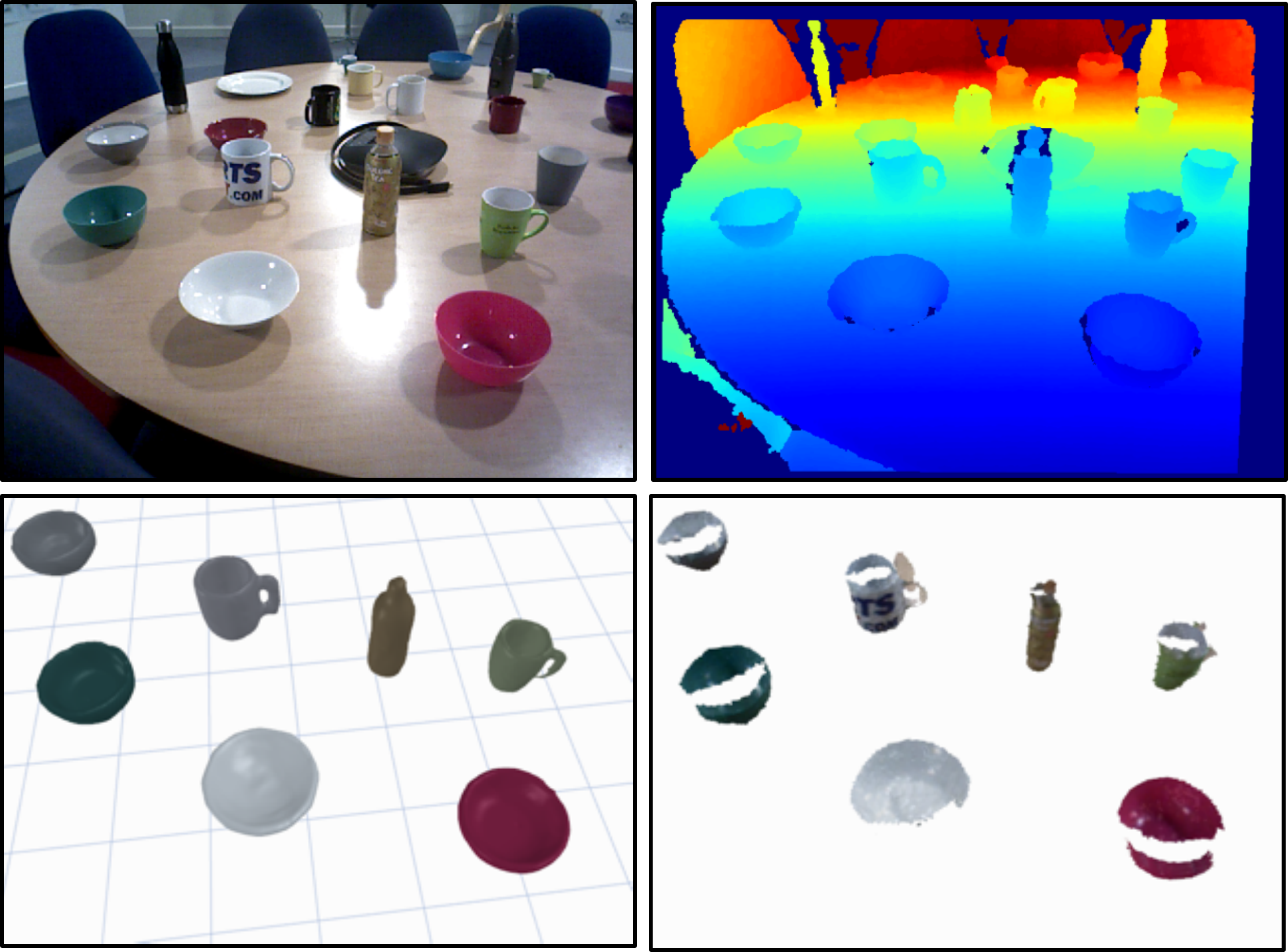}
\end{center}
\caption{\textbf{Top:} Compact, optimisable shape models used in an object-level SLAM system which maps a real world cluttered table top scene with varied object shapes from different classes.  \textbf{Bottom:} Class-level priors allow accurate and complete object reconstruction (bottom-left) even from a single image in contrast to partial reconstruction from TSDF fusion (bottom-right).}
\label{fig:teaser}
\vspace{2mm} \hrule \vspace{-2em}
\end{figure}

There have been two main approaches for 3D shape reconstruction from images. Classical reconstruction techniques infer geometry by minimizing the discrepancy between a reconstructed 3D model and observed data through a measurement function \cite{Izadi:etal:UIST2011,McCormac:etal:3DV2018,Whelan:etal:RSS2015}.  These methods are flexible and general, but they can only reconstruct directly observed parts of a scene and are limited in accuracy when observations are weak or noisy. On the other hand, discriminative methods learn to map image measurements to 3D shape, such as through a feed-forward neural network \cite{Gkioxari:etal:ICCV2019,Kundu:etal:CVPR2018,Wu:etal:NIPS2017,Tulsiani:etal:CVPR2017,Wang:etal:ECCV2018,Wu:etal:CVPR2015}. These methods take advantage of regularities in data for robustness but have trouble in generalisation and lack the ability to integrate multiple measurements in a principled way. 

Our work sits between these two approaches. We capture regularities in data though a volumetric 3D generative model represented though a class conditioned Variational Auto Encoder (VAE), trained on a collection of CAD models, allowing us  to represent object shape through a compact code. We then use the generative model for shape inference though iterative optimisation of the latent code with respect to any number of depth image measurements. 

To use a  generative method for inference we need a rendering function to transform 3D volumes into measurements; in our case depth images with object segmentation. The design of this function will influence optimisation speed and convergence success. Two important design considerations are (1) receptive field, the size of 3D region which influences each rendered pixel, and (2) uncertainty modeling, the confidence of each rendered pixel depth. We introduce a novel probabilistic volumetric rendering function based on these two design principles, 
improving the state of the art in volumetric rendering.

In scenes with many objects, our optimisable compact object models can serve as the landmarks in a SLAM system, where we use the same measurement function for camera tracking, object poses and shape optimisation. 
We quantitatively show that joint optimisation leads to more robust tracking and reconstruction, with comparable surface reconstruction to the data driven Fusion++ \cite{McCormac:etal:3DV2018}, while reaching full object reconstruction from far fewer observations.

An emphasis of this paper is to design object models that work robustly in the real world.  We demonstrate the robustness of our proposed rendering function through qualitative demonstrations of our object-level SLAM on real world image sequences from a cluttered table-top scene obtained with a noisy depth camera, and on an augmented reality demo. Furthermore we integrate our efficient shape inference method into a real time robotic system, and show that the completeness and accuracy of our object reconstructions enable robotic tasks such as packing objects into a tight box or sorting objects by shape size. We encourage readers to watch the associated \textbf{video} which supports our submission. 

To summarise, the key contributions of our paper are: (i) A novel volumetric probabilistic rendering function which enables robust and efficient multi-view shape optimisation.
    (ii) The first object-level SLAM capable of jointly optimising full object shapes and poses together with camera trajectory from real world images. 
    (iii) The integration into a real-time robotic system that can achieve useful manipulation tasks with varied object shapes from different categories due to complete high quality surface reconstructions.

\section{Related Work}
There are two categories of methods for full object shape inference from images, discriminative and generative approaches. Discriminative single image reconstruction approaches such as \cite{Gkioxari:etal:ICCV2019,Kundu:etal:CVPR2018,Wu:etal:NIPS2017,Tulsiani:etal:CVPR2017,Wang:etal:ECCV2018,Wu:etal:CVPR2015} lack the ability to integrate multiple observations in a principled way. For example, DeepSLAM++ \cite{Hu:etal:ARXIV2019} directly averages 3D shapes predicted by Pix3D \cite{Sun:etal:CVPR2018}, while 3D-R2N2 \cite{Choy:etal:ECCV2016} uses a recurrent network to update shapes from new images.

Generative methods for shape reconstruction were first developed using linear models such as PCA \cite{Amaury:etal:CVPR2013,Engelmann:etal:WACV2017,Wang:etal:ARXIV2019,Li:etal:BMVC2019,Zhu:etal:WACV2018} which constrained them to simple shapes. Our work comes at a time when many authors are building accurate and flexible generative models for objects using neural networks \cite{Jiajun:etal:NIPS2016,Park:etal:CVPR2019,Mescheder:etal:CVPR2019,Paschalidou:etal:CVPR2019} and starting to us them in shape inference from images \cite{Li:etal:CVPR2020, Jiang:etal:CVPR2020}. However existing methods for generative full shape inference have not yet shown to work in incremental real-world multi-object settings in real time.

To allow for robust inference, critical in real world settings, and efficient optimisation, necessary for real time application, we introduce a novel differential probabilistic rendering formulation. Existing differential volumetric rendering techniques used for shape inference such as \cite{Liu:etal:CVPR2020, Niemeyer:etal:CVPR2020, Jiang:etal:CVPR2020} have a local receptive field, which means each rendered pixel only depends on a single 3D point sample; this causes optimisation to be slow and easily stuck in local minima because of local gradient flow. We take inspiration from traditional volumetric rendering in graphics \cite{Kajiya:etal:1984} to define a probabilistic rendering formulation, which gathers occupancy probabilities along samples of a back-projected ray for each pixel to render a depth image. Additionally, we build a Gaussian pyramid after rendering to increase the spatial receptive field.  Our probabilistic formulation allows us to measure the uncertainty of each rendered pixel which improves optimisation by weighting residuals. 

One of the targets of visual SLAM research has been  a continuous improvement in the richness of scene representations, from
sparse SLAM systems which reconstruct point clouds, \cite{Davison:etal:PAMI2007,Engel:etal:PAMI2017}
via dense surface representations \cite{Newcombe:etal:ISMAR2011, Whelan:etal:RSS2015} to semantically labelled dense maps \cite{McCormac:etal:ICRA2017,  Xiang:Fox:ARXIV2017}. Dense maps of whole scenes are very expensive to store, and difficult to optimise probabilistically from multiple views. A sensible route towards representations which are both efficient and semantically meaningful is to focus  representation  resources  on the most important elements of a scene, objects. 

Dense object-based SLAM has been previously attempted, but our work fills a gap between systems which make separate reconstructions of every identified object
\cite{McCormac:etal:3DV2018,Sunderhauf:etal:IROS2017}, which are general with respect to arbitrary classes but can only reconstruct directly observed surfaces, and those which populate maps with instances of object CAD models \cite{Salas-Moreno:etal:CVPR2013}, which are efficient but limited to precisely known objects. Our approach can efficiently reconstruct whole objects which are recognised at the class level, and cope with a good range of shape variation within each class.

\section{Class-Level Object Shape Descriptors}
\label{sec:shape}
Objects of the same semantic class exhibit strong regularities in shape under a common pose alignment. We leverage this to construct the class specific smooth latent space, which allows us to represent the shape of an instance with a small number of parameters by training a single Class-Conditional Variational Autoencoder neural network.

\begin{figure}[tbp]
\begin{center}
   \includegraphics[width=0.9\linewidth]{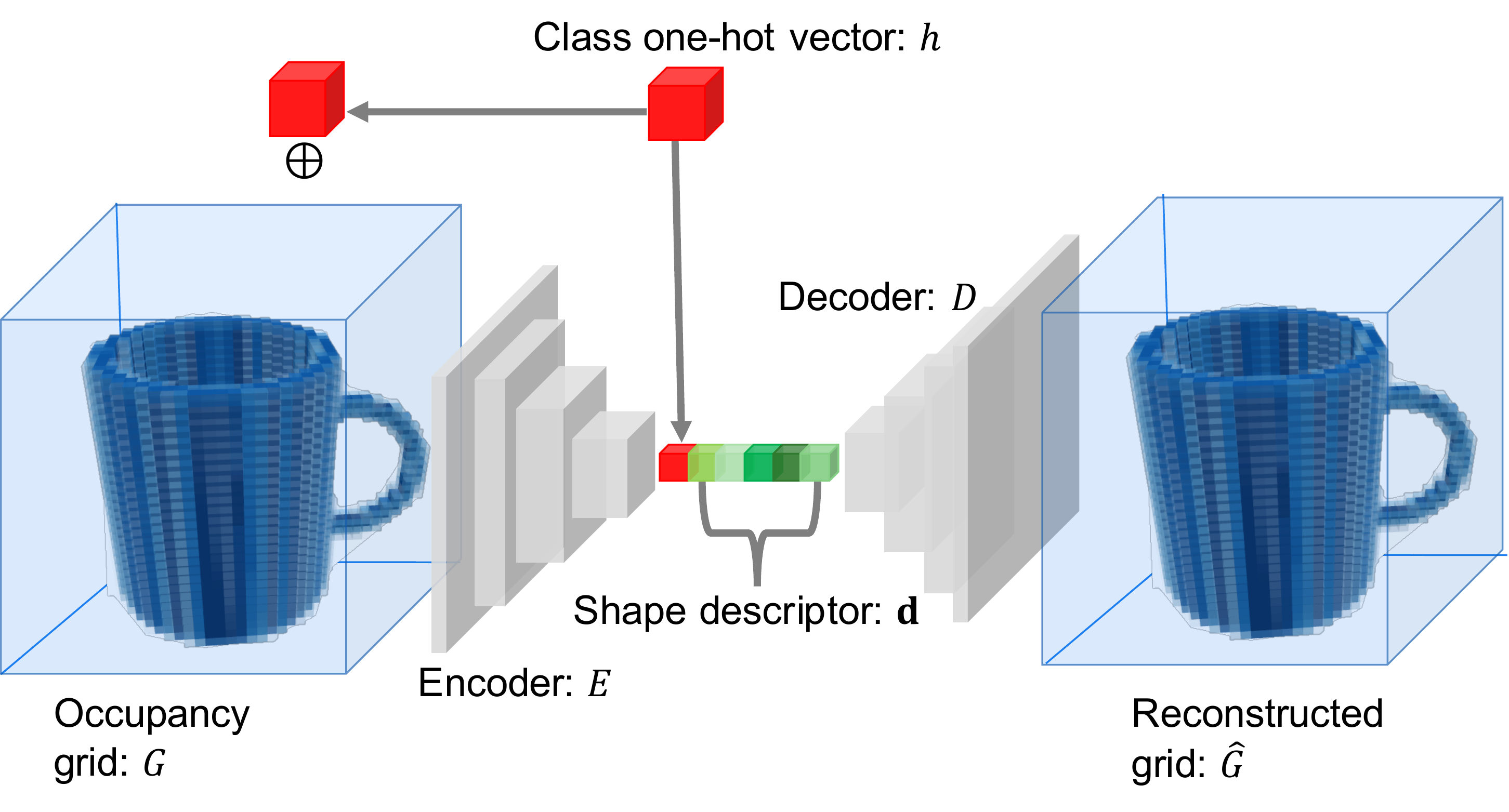}
\end{center}
   \caption{\textbf{Occupancy Variational Autoencoder}: The class one hot vector $h$ is concatenated channel-wise to each occupancy voxel in the input occupancy grid $G$. The input is compressed into shape descriptor $\mathbf{d}$ by encoder network $E$. The shape descriptor and the class-one hot vector are concatenated and passed through decoder network $D$ to obtain occupancy reconstruction $\hat{G}$.}
\label{fig:vae}
    \vspace{2mm} \hrule
\end{figure}

\subsection{Network Design}
3D object shapes are represented by voxel occupancy grids of dimension $32 \times 32 \times 32$, with each voxel storing a continuous occupancy probability between 0 and 1. A voxel grid was chosen to enable shapes of arbitrary topology.

The 3D models were obtained from the ShapeNet database \cite{Shapenet:ARXIV2015}, which comes with annotated model alignment. The occupancy grids were obtained by converting the model meshes into a high resolution binary occupancy grid, and then down-sampling by average pooling.

A single 3D CNN Variational Autoencoder \cite{Kingma:Welling:ICLR2014} was trained on objects from 4 classes: `mug', `bowl', `bottle', and `can', common table-top items. The encoder is conditioned on the class by concatenating the class one-hot vector as an extra channel to each occupancy voxel in the input, while the decoder is conditioned by concatenating the class one-hot vector to the encoded shape descriptor, similar to \cite{Sohn:etal:NIPS2015,Tan:etal:CVPR2018}. A KL-divergence loss is used in the latent shape space, while a binary-crossentropy loss is used for reconstruction. We choose a latent shape variable of size 16. The 3D CNN encoder has 5 convolutional layers with kernel size 4 and stride 2, each layer doubles the channel size except the first one which increases it to 16. The decoder mirrors the encoder using deconvolutions.

The elements of our VAE network are shown in Figure~\ref{fig:vae}.

\section{Probabilistic Rendering}
\label{sec:render}
Rendering is the process of projecting a 3D model into image space. Given the pose of the grid with respect to the camera $T_{CG}$, we wish to render a depth image. We denote the rendered depth image as $\hat{\delta}_{\mu}$ with uncertainty $\hat{\delta}_{var}$, and the rendering function $R()$, such that $\hat{\delta}_{\mu}, \hat{\delta}_{var}=R(G, T_{GC})$.
When designing our render function, we wish for it to satisfy three important requirements: to be differentiable and probabilistic so that it can be used for principled inference, and to have a wide receptive field so that its gradients behave properly during optimisation. These features make a robust function that can handle real world noisy measurements such as depth images.

We now describe the algorithm for obtaining the depth value for pixel $(u,v)$:

\vspace{-1.5em}
\paragraph{\textbf{Point sampling.}} Sample $M$ points uniformly along backprojected ray $r$, in depth range $[\hat{\delta}_{min}, \hat{\delta}_{max}]$. Each sampled depth $\hat{\delta}_i = \hat{\delta}_{min}+\frac{i}{M}(\hat{\delta}_{max}-\hat{\delta}_{min})$ and position in the camera frame $s_i^C = \hat{\delta}_i r$. Each sampled point is transformed into the voxel grid coordinate frame as $s_i^G = T_{GC} s_i^C$.
    
\vspace{-1.5em}
\paragraph{\textbf{Occupancy interpolation.}} Obtain occupancy probability $o_i=Tril(s_i^O, G)$, for point $s_i^O$ from the occupancy grid, using trilinear interpolation from its 8 neighbouring voxels. 
    
\begin{figure}[b]
\begin{center}
   \includegraphics[width=0.9 \linewidth]{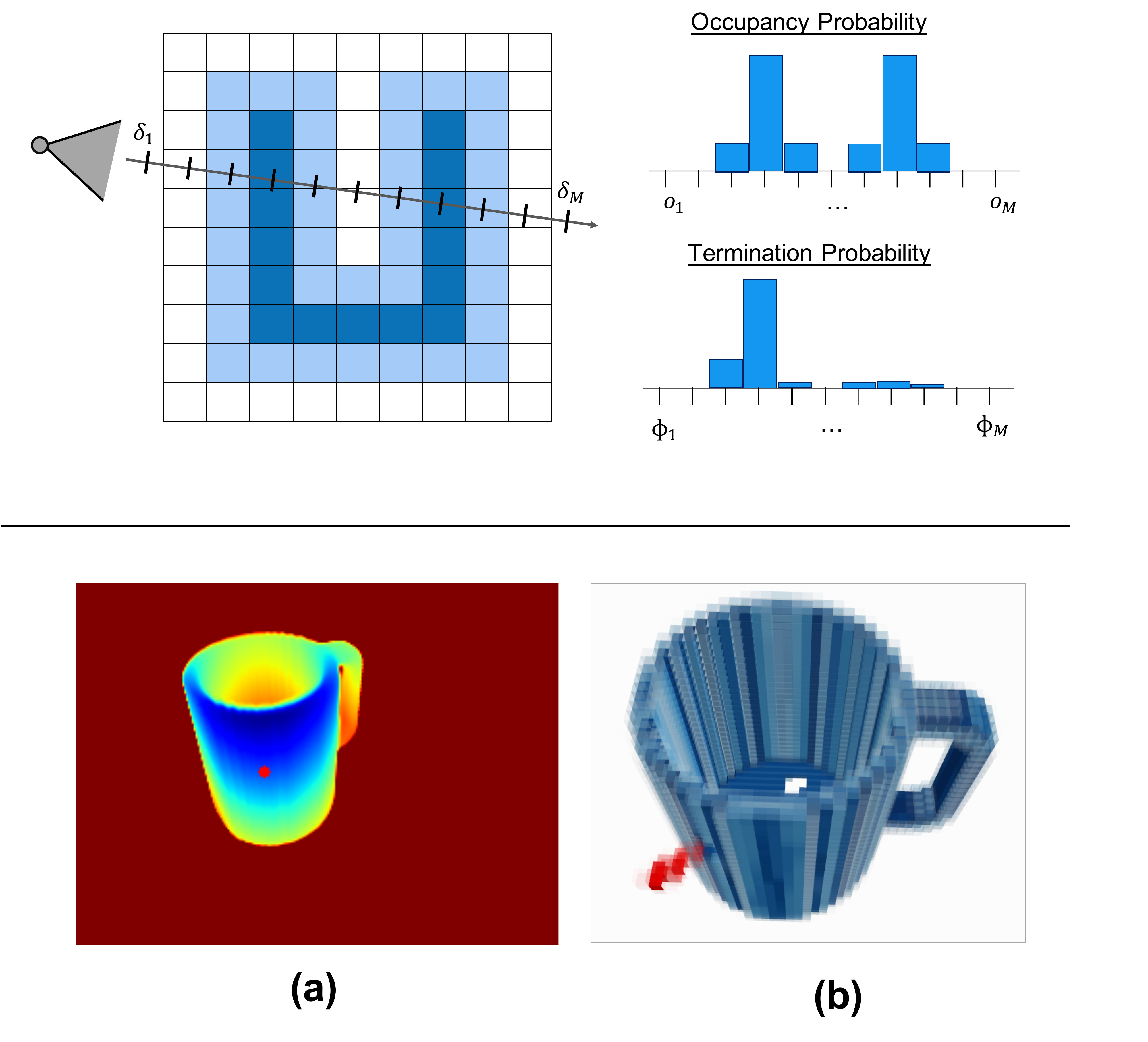}
\end{center}
   \caption{\textbf{Pixel rendering}: Each pixel is back-projected into a ray from which uniform depth samples $\delta_i$ are taken. Occupancy probability $o_i$ is obtained from the voxel grid by trilinear interpolation, and termination probability $\phi_i$ is calculated. \textbf{(a):} A $32\times32\times32$ mug occupancy grid. \textbf{(b):} The derivative of the highlighted red pixel with respect to occupancy values is shown in red.}
\label{fig:ray}
    \vspace{2mm} \hrule
\end{figure}

\vspace{-1.5em}
\paragraph{\textbf{Termination probability.}} We denote the depth at pixel  $[u,v]$ by $\mathbf{D}[u,v]$. Now we can calculate $p(\mathbf{D}[u,v]=\hat{\delta}_i)$ (that is, the termination probability at depth $\hat{\delta}_i$) as:
\begin{equation}
        \phi_i = p(\mathbf{D}[u,v]=\hat{\delta}_i) =  o_i \prod_{j=1}^{i-1}(1-o_j)~.
\end{equation}
Figure~\ref{fig:ray} relates occupancy and termination probabilities.
   
\vspace{-1.5em}
\paragraph{\textbf{Escape probability}} Now we define the escape probability (the probability that the ray doesn't intersect the object) as:
\begin{equation}
        \phi_{M+1} = p(\mathbf{D}[u,v]>\hat{\delta}_{max}) = \prod_{j=1}^{M}(1-o_j)~,
\end{equation} $\{\phi_i\}$ forms a discrete probability distribution.

\vspace{-1.5em}
\paragraph{\textbf{Aggregation}} We obtain the rendered depth at pixel $[u,v]$ as the expected value of the random variable $\mathbf{D}[u,v]$:
\begin{equation}
            \hat{\delta}_{\mu}[u,v] = \mathbb{E}[\mathbf{D}[u,v]] = \sum_{i=1}^{M+1} \phi_i \hat{\delta}_i~.
\end{equation}
$d_{M+1}$, the depth associated to the escape probability is set to $1.1 d_{max}$ for practical reasons.

\vspace{-1.5em}
\paragraph{\textbf{Uncertainty}}  Depth uncertainty is calculated as:
\begin{equation}
        \hat{\delta}_{var}[u,v] = Var[\mathbf{D}[u,v]] = \sum_{i=1}^{M+1} \phi_i (\hat{\delta}_i-D[u,v])^2~.
\end{equation}

\vspace{-1.5em}
\paragraph{\textbf{Mask}} Note that we can render a segmentation mask as:
 \begin{equation}
 m[u,v]=1-\phi_{M+1}
\end{equation}
        
For multi-object rendering we combine all the renders by taking the minimum depth at each pixel, to deal with cases when objects occlude each other: 
\begin{equation}
\begin{split}
    \hat{\delta}_{\mu}[u,v] &= R(\{\hat{G}_i\}, \{T_{G}^i\}, T_C)[u,v] \\
    &= min\{\hat{\delta}_{\mu}^1[u,v], ..., \hat{\delta}_{\mu}^N[u,v]\}
    ~.
\end{split}
\end{equation}

Figure \ref{fig:ray} shows the relation between rendered depth and occupancy probabilities. Additionally, we apply Gaussian blur down-sampling to the resulting rendered image at different pyramid levels (4 levels with 1 pixel standard deviation each) to perform coarse to fine optimisation, this increases the spatial receptive field in the higher levels of the pyramid because each rendered pixel is associated to several back projected rays.
\section{Object Shape and Pose Inference}
\label{sec:inference}
\begin{figure*}[htbp]
\begin{center}
   \includegraphics[width=0.8\linewidth]{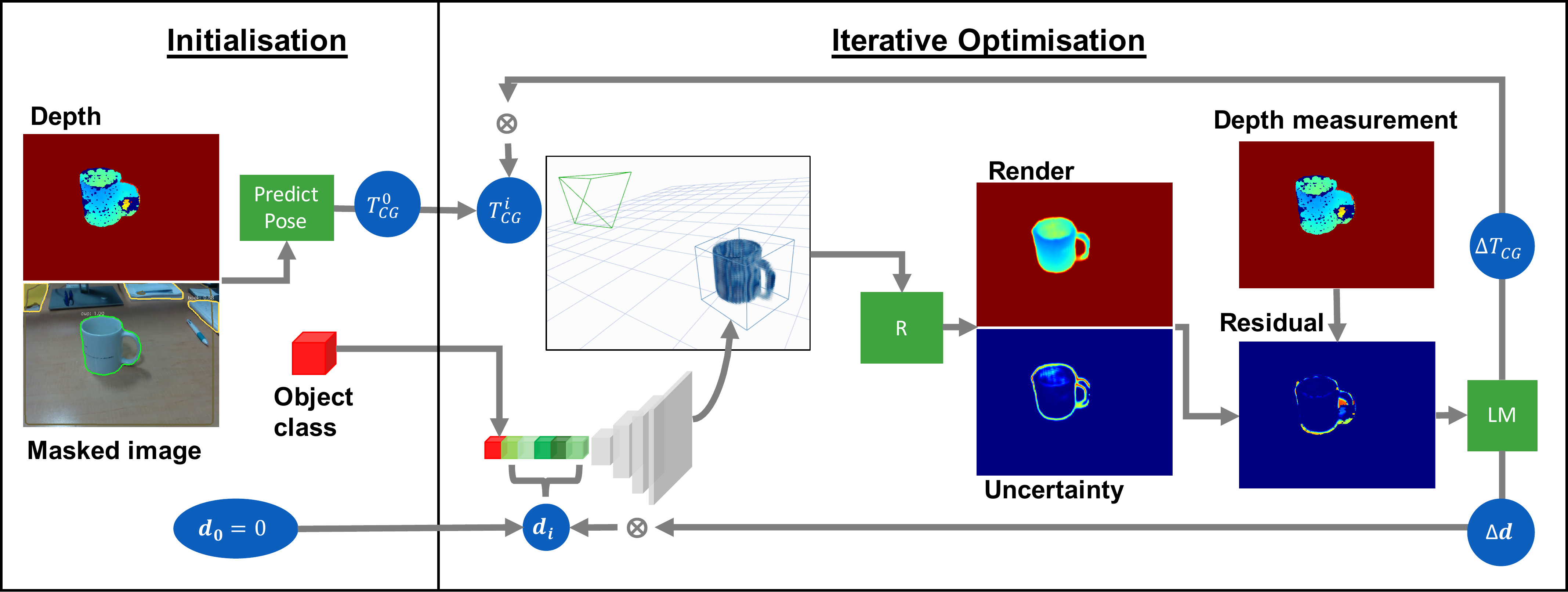}
\end{center}
   \caption{\textbf{Initialisation:} Initial object pose $T_{CG}^0$ is estimated from a depth image and masked RGB image; object class is inferred from RGB only. The shape descriptor $\mathbf{d}$ is set to 0, representing the mean class shape.  \textbf{Optimisation:} The shape descriptor is decoded into a full voxel grid, which is used with the pose to render an object depth map. The least squares residual between this and depth is used  update the shape descriptor and object pose iteratively with the Levenberg-Marquardt algorithm.}
\label{fig:inference}
    \vspace{2mm} \hrule 
\end{figure*}

Given a depth image from an object of a known class, we wish to infer the full shape and pose of the object. We assume we have a segmentation mask and classification of the object, which in our case is obtained with Mask-RCNN \cite{He:etal:ICCV2017}.
To formulate our inference method, we integrate the object shape models developed on Section~\ref{sec:shape} with a measurement function, the probabilistic render algorithm outlined in Section~\ref{sec:render}. We will now describe the inference algorithm for a single object observation setup, and this will be extended to multiple objects and multiple observations in the SLAM system described in Section~\ref{sec:SLAM}.

\subsection{Shape and Pose Optimisation}
An object's pose $T_{CG}$ is represented as a 9-DoF homogeneous transform with $\mathcal{R}$, $t$, and $S$ the rotation, translation and scale of the object with respect to the camera.

The shape of the object is represented with latent code $\mathbf{d}$, which is decoded into full occupancy grid $\hat{G}$ using the decoder described in Section \ref{sec:shape}.

We wish to find the pose and shape parameters that best explain our depth measurement $\delta$. We consider the rendering  $\mathbf{D}$ of the object as Gaussian distributed, with mean $\hat{\delta}_{\mu}$ and variance $\hat{\delta}_{var}$ calculated through the render function:
\begin{equation}
\begin{split}
\hat{\delta}_{\mu}, \hat{\delta}_{var} &= R(\hat{G}, T_{GC}) \\
&= R(D(\mathbf{d},h), T_{GC}) 
~,
\end{split}
\end{equation} with $h$ the class one-hot vector of the detected object.

When training the latent shape space a Gaussian prior distribution is assumed on the shape descriptor. With this assumption and by taking $\hat{\delta}_{var}$ as constant, our MAP objective takes the form of least squares problem. We apply the Levenberg–Marquardt algorithm for estimation:

\begin{equation}
\label{eq:infer}
\begin{split}
    &\min \limits_{\mathbf{d}, T_{CG}} -log(p(\delta | \mathbf{d}, T_{CG} ) p(\mathbf{d})) \\
    &=
    \min \limits_{\mathbf{d}, T_{CG}} (L_{render}(\mathbf{d}, T_{CG})+L_{prior}(\mathbf{d}))\\
    &=
    \min \limits_{\mathbf{d}, T_{CG}} (\sum_{u,v} \frac{(\delta [u,v] - \hat{\delta}_{\mu}[u,v])^2}{ \hat{\delta}_{var}[u,v]} + \sum_{i} \mathbf{d}_i^2).
\end{split}
\end{equation}
A structural prior is added to the optimisation loss to force the bottom on the object to be in contact with the supporting plane. We render an image from a virtual camera under the object and recover the surface mesh from the occupancy grid by marching cubes. Figure \ref{fig:inference} illustrates the single object shape and pose inference pipeline.

\subsection{Variable Initialisation}

Second order optimisation methods such as Levenberg–Marquardt require a good initialisation.  The object's translation and scale are intitialised using the backprojected point cloud from the masked depth image. The first is set to the centroid of the point cloud, while the latter is recovered from the centroid's distance to the point cloud boundary.

Our model classes (`mug', `bowl', `bottle', and `can') are often found in a vertical orientation in a horizontal surface. For this reason we detect the horizontal surface using the point cloud from the depth image and initialise the object's orientation to be parallel to the surface normal. One class of objects, `mug', is not symmetric around the vertical axis. To initialise the rotation of this class along the vertical axis we train a CNN. The network takes as input the cropped object from the RGB image, and outputs a single rotation angle along the vertical object axis. We train the network in simulation using realistic renders from a physically-based rendering engine, Pyrender, randomising object's material, lighting positions and colors. The network has a VGG-11 \cite{Simonyan:Zisserman:ICLR2015} backbone pre-trained on ImageNet \cite{Russakovsky:etal:ILSVRC15}.

\begin{figure}[b]
\begin{center}
   \includegraphics[width=0.8\linewidth]{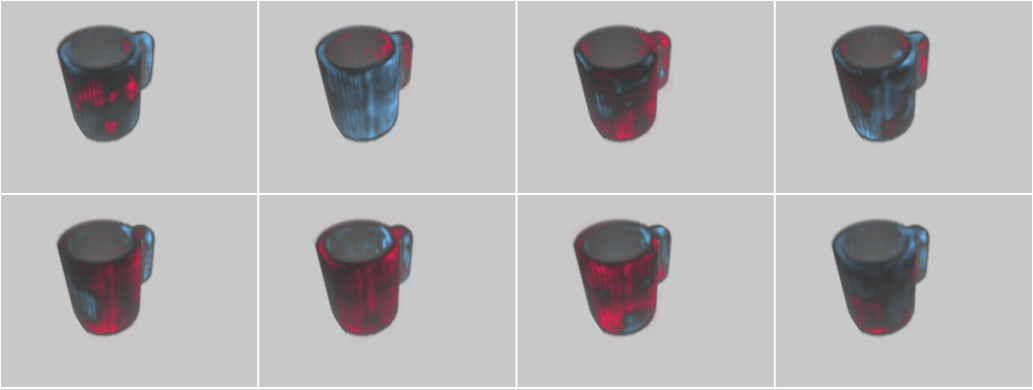}
\end{center}
   \caption{\textbf{Shape descriptor influence:} the derivative of a rendered decoded voxel grid with respect to 8 entries of the shape descriptor.}
\label{fig:code_deriv}
    \vspace{2mm} \hrule
\end{figure}

The shape descriptor is initialised to $\mathbf{d}=0$, which gives the mean class shape under the Gaussian prior of a VAE. Optimisation iteratively deforms the mean shape to best fit our observations. Figure \ref{fig:code_deriv} illustrates how changes in the shape descriptor alter the shape of the object.  

\section{Object-Level SLAM System}
\label{sec:SLAM}
We have developed class level shape models and a measurement function that allows us to infer object shape and pose from a single RGB-D image.
From stream of images we want to incrementally build a map of all the objects in a scene while simultaneously tracking the position of the camera. For this,  we will show how to use the render module for camera tracking, and for joint optimisation of camera poses, object shapes, and object poses with respect to multiple image measurements. This will allow us to construct a full, incremental, jointly optimisable object-level SLAM system with sliding keyframe window optimisation. 

\subsection{Data association and Object Initialisation}
\label{subsec:association}
For each incoming image, we first segment and detect the classes of all objects in the image using Mask-RCNN \cite{He:etal:ICCV2017}. For each detected object instance, we try to associate it with one of the objects already reconstructed in the map. This is done in a two stage process:

\textbf{Previous frame matching}: We match the masks in the image with masks from the previous frame. Two segmentations are considered a match if their IoU is above 0.2.

\textbf{Object mask rendering}: If a mask is not matched in stage 1, we try to match it directly with map objects by rendering their masks and computing IoU overlaps.

If a segmentation is not matched with any existing objects we initialise a new object as in Section~\ref{sec:inference}.

\subsection{Camera Tracking}

We wish to track the camera pose $T_C^j$ for the latest depth measurement $\delta_j$. Once we have performed association between segmentation masks and reconstructed objects as described in Section~\ref{subsec:association}, we have a list of matched object descriptors $\{ \mathbf{d_1}, ..., \mathbf{d_N} \}$. We initialise our estimate for $T_C^j$ as the tracked pose of the previous frame $T_C^{j-1}$, and render the matched objects as described in Section~\ref{sec:render}:
\begin{equation}
    \hat{\delta}_{\mu}, \hat{\delta}_{var} = R(\{\hat{G}_i\}, \{T_{G}^i\}, T_C^j)
~.
\end{equation}
The loss between render and measured depth is:
\begin{equation}
\label{eq:lrender}
   L_{render}(\{\mathbf{d}_i\}, \{T_{G}^i\}, T_C^j) = \sum_{u,v} \frac{(\delta_j [u,v] - \hat{\delta}_{\mu}[u,v])^2}{ \hat{\delta}_{var}[u,v]}
   ~.
\end{equation}
Notice that this is the same loss used when inferring object pose and shape, but now we assume that the map (the object shapes and poses) is fixed and we want to estimate the camera pose $T_C^j$. As before, we use the iterative Levenberg–Marquardt optimisation algorithm.

\subsection{Sliding-Window Joint Optimisation}

\begin{figure}[t]
\begin{center}
   \includegraphics[width=0.9\linewidth]{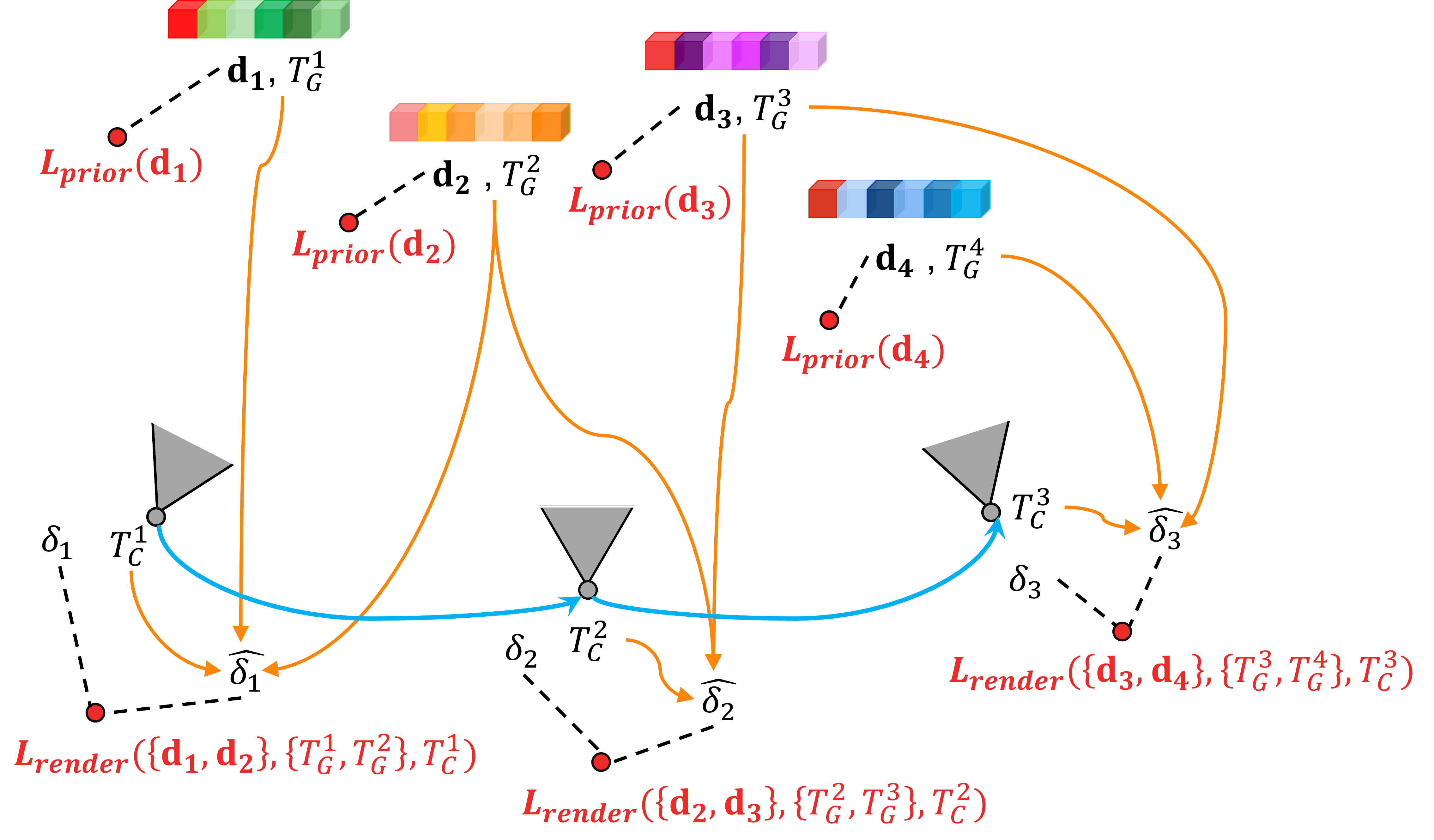}
\end{center}
   \caption{\textbf{Optimisation graph}, showing all jointly-optimised variables. Render and prior factors connect the different variables. A render factor compares object shape renders with depth measurements. Prior factors constrain how much each object shape can deviate from the mean shape of its class.}
\label{fig:joint}
    \vspace{2mm} \hrule
\end{figure}

We have shown how to reconstruct objects from a single observation, and how to track the position of the camera by assuming the map is fixed. This will however lead to the accumulation of errors, causing motion drift. Integrating new viewpoint observations for an object is also desirable, to improve its shape reconstruction.  To tackle these two challenges, we wish to jointly optimise a bundle of camera poses, object poses, and object shapes. Doing this with all frames is however computationally infeasible, so we jointly optimise the variables associated to a select group of frames, called keyframes, in a sliding window manner, following the philosophy introduced by PTAM \cite{Klein:Murray:ISMAR2007}. 

\textbf{Keyframe criteria:}
There are two criteria for selecting a frame as a keyframe. If an object was initialised in the frame then it is selected as a keyframe, or second if the frame viewpoint for any of the existing objects is larger than 13 degrees from the frame in which the object was initialised.

\textbf{Bundle Optimisation:}
Each time that a frame is selected as a keyframe we jointly optimise the variables associated with a bundle of $N$ keyframes. In particular we select a window of 3 keyframes, the new keyframe and its two closest keyframes, with the previously defined distance.

To formulate the joint optimisation loss, consider, $T_C^1$, $T_C^2$, and $T_C^3$, the poses of the keyframes in the optimisation window; $T_C^1$ is held fixed. Now suppose $\{ \mathbf{d}_i \}$ is the set of shape descriptors for the objects observed by the three keyframes. Then we can render a depth image and uncertainty for each keyframe as:
\begin{equation}
    \hat{\delta}_{\mu}^j, \hat{\delta}_{var}^j = R(\{\hat{G}_i\}, \{T_{G}^i\}, T_C^j)
~,
\end{equation} 
with $\hat{G}_i=D(\mathbf{d}_i, h_i)$.
For each render we compute a loss with the respective depth measurement, $L_{render}^j$ as in Equation~\ref{eq:lrender}, and a prior loss, $L_{prior}^i$ on all codes as in Equation~\ref{eq:infer}. Figure \ref{fig:joint} illustrates the joint optimisation problem. Our final loss, optimised using Levenberg-Marquardt, is:
\begin{equation}
\begin{split}
   &L_{joint}(\{\mathbf{d}_i\}, \{T_{G}^i\}, \{T_C^j\}) =\\
   & \sum_j L_{render}(\{\mathbf{d}_i\}, \{T_{G}^i\}, T_C^j) + \sum_i L_{prior}(\mathbf{d}_i)
   ~.
\end{split}
\end{equation}

\textbf{Timings:} Rendering a single object takes 7ms, full object reconstruction takes approximately 1.5 seconds. Camera tracking is 7fps and joint optimisation takes 2 seconds.

\begin{figure}[htbp]
\begin{center}
\includegraphics[width=0.7\linewidth]{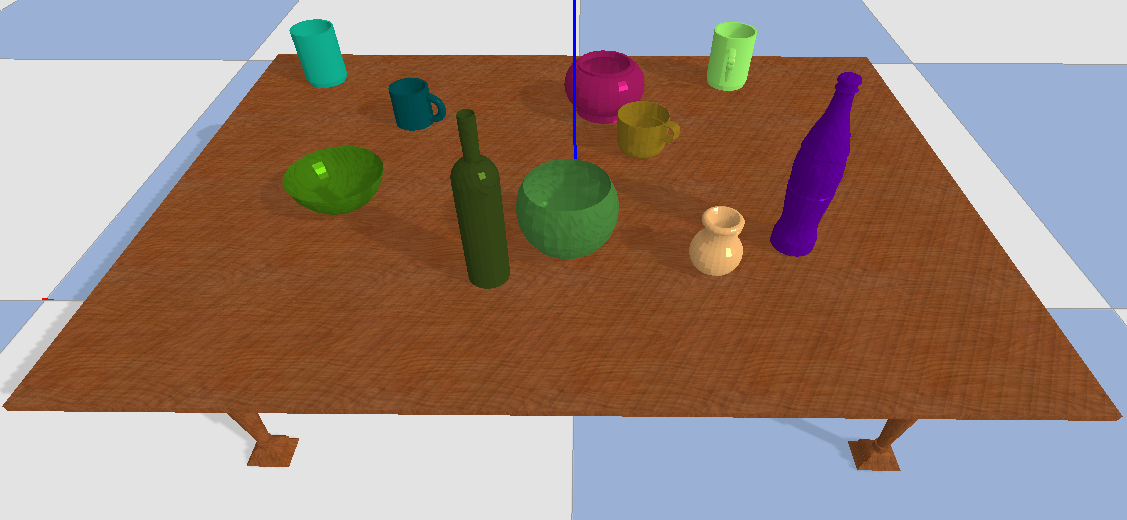}
\includegraphics[width=0.7\linewidth]{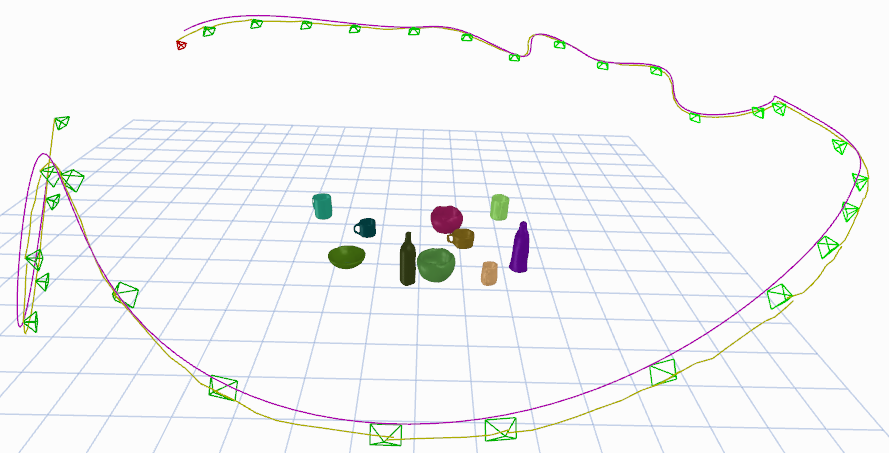}
\end{center}
\caption{Synthetic scene example along with reconstruction and camera trajectory. Ground truth trajectory is shown in purple and tracked one in yellow, keyframes with green frustum.}
\label{fig:synthetic}
    \vspace{2mm} \hrule
\end{figure}

\section{Experimental Results}
\label{sec:eval}

\subsection{Metrics}

For shape reconstruction evaluation we use three metrics: chamfer-$L_1$ distance and accuracy as defined in \cite{Mescheder:etal:CVPR2019} and completeness (with 1cm threshold) as defined in \cite{Li:etal:CVPR2020}. We sample 20000 points on both reconstruction and ground truth CAD model meshes.

\subsection{Rendering Evaluation}
\label{sec:render_eval}

\begin{table}[h!]
  \begin{center}
    \caption{Shape reconstruction results for 1, 2, and 3 views. We do an ablation study of our method and compare with DVR \cite{Niemeyer:etal:CVPR2020}.}
    \label{tab:table_render}
    \scriptsize
    \begin{tabular}{c|c|c|c|c|c}
      \toprule 
       & \textbf{Full} & \textbf{No Unc.} & \textbf{No Gauss.} & \textbf{ \cite{Niemeyer:etal:CVPR2020}} & \textbf{Mask}\\
      \midrule 
      \textbf{\underline{1 view}} & & & & & \\
      accuracy [\textit{mm}] & 4.459 & 4.998 & 4.701 & 8.967 & 15.806\\
      chamfer-$L_1$ [\textit{mm}] & 4.439 &4.844 & 4.928 & 11.896 & 18.386\\
      completion [1cm] & 93.492 & 91.857 & 90.812 & 43.075 & 30.212\\
      \midrule 
      \textbf{\underline{2 views}} & & & & \\
      accuracy [\textit{mm}] & 3.752 &4.270 & 4.237 & 8.408 & 4.709\\
      chamfer-$L_1$ [\textit{mm}] & 3.854 & 4.185 & 4.723 & 11.325 & 4.438\\
      completion [1cm] & 95.72 & 94.627 & 90.752 & 43.342 & 93.73\\
      \midrule 
      \textbf{\underline{3 views}} & & & & \\
      accuracy [\textit{mm}] & 3.484 & 4.158 & 3.827 & 8.277 & 4.620\\
      chamfer-$L_1$ [\textit{mm}] & 3.648 & 4.010 & 4.281 & 10.913 & 4.210\\
      completion [1cm] & 96.065 & 95.165 & 93 & 44.815 & 95.44\\
      \bottomrule 
    \end{tabular}
  \end{center}
\end{table}

In this evaluation test the optimisation performance of our rendering formulation. We perform object shape and pose optimisation on all the objects of the `mug' category in the ShapeNet dataset. For each instance we generate three random views of the object. Initial object pose is predicted from the first view. We perform 30 optimisation iterations for 1, 2, and 3 views. Table \ref{tab:table_render} shows median shape accuracy, completion, and chamfer distance after optimisation. We compare our full system with versions without uncertainty, without Gaussian pyramid, and with a loss only between the rendered and Mask-RCNN segmentation masks. We compare with the state of the art volumetric differential rendering component in paper Differential Volumetric Rendering (DVR) \cite{Niemeyer:etal:CVPR2020} with our shape representation.

We observe that additional views improve shape reconstruction, more drastically in the mask optimisation because of the scale ambiguity in a single image. We also see that both the uncertainty and Gaussian pyramid are necessary for more accurate and complete shape reconstructions. Our method significantly improves on DVR, which is both less precise and has much lower shape completion, because of its local receptive field.

\begin{figure}[t]
\begin{center}
   \includegraphics[width=0.8\linewidth]{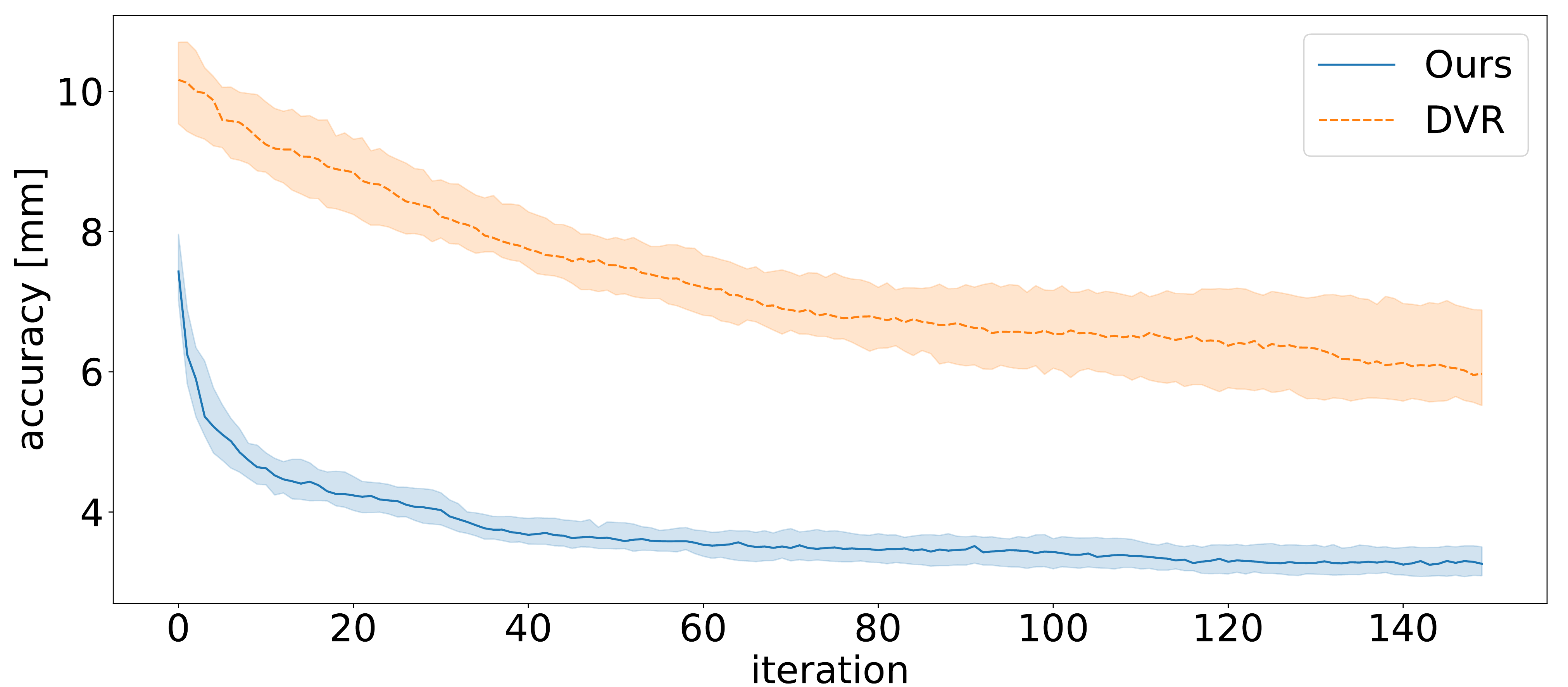}
\end{center}
   \caption{Median reconstruction accuracy (95\% confidence) across 150 optimisation iterations of all `mug' objects instances comparing our proposed renderer with \cite{Niemeyer:etal:CVPR2020}.}
\label{fig:convergence}
\end{figure}

To further illustrate the comparison, we plot in Figure \ref{fig:convergence} median reconstruction accuracy across 150 optimisation iterations with all object instances against our proposed method. The plot illustrates the much faster convergence of our method and its ability to reach a lower accuracy error.

\subsection{SLAM evauation} 
\label{sec:slam_eval}

In this evaluation we evaluate our full SLAM system and how it generalises to new object instances. We create a synthetic dataset. Random object CAD models are spawned on top of a table model with random positions and vertical orientation. Five scenes are created with 10 different objects on each from three classes: `mug', `bowl', and `bottle'. The models are obtained from the ModelNet40 dataset \cite{Wu:etal:CVPR2015} which are not used during training of the shape model.  

For each scene a random trajectory is generated by sampling and interpolating random camera positions and look at points in the volume bounded by the table. Image and depth renders are obtained from the trajectory with PyBullet render, which is different rendering engine than the one used for training pose prediction. 

\subsubsection{Fusion++ comparison}

\begin{figure}[bp]
\begin{center}
\includegraphics[width=0.45\linewidth]{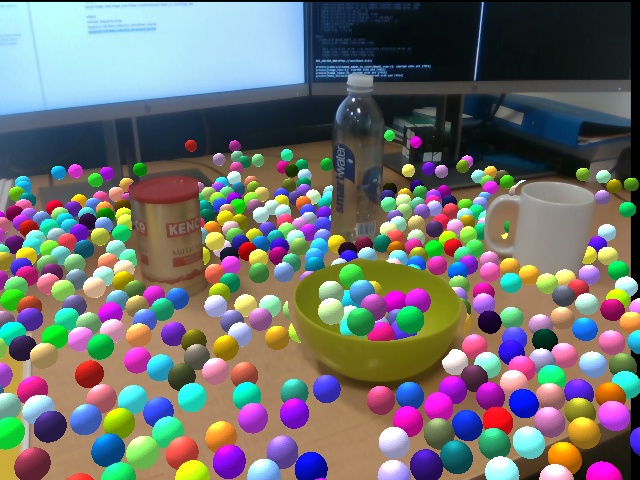}
\includegraphics[width=0.5\linewidth]{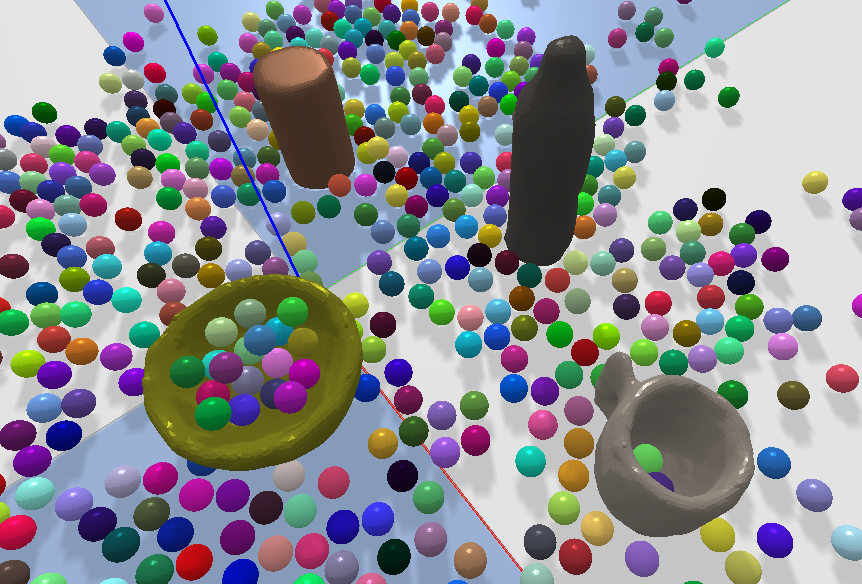}
\end{center}
\caption{\textbf{Few-shot augmented reality:} Complete and watertight meshes can be obtained from few images due to the learned shape priors. This are then loaded into a physics engine to perform realistic augmented reality demonstrations.}
\label{fig:ar}
    \vspace{2mm} \hrule
\end{figure}

We compare our proposed method with a custom implementation of Fusion++ \cite{McCormac:etal:3DV2018} using open-source TSDF fusion \cite{Yi-Zhou:etal:ARXIV2018} for each object volume. In this experiment ground truth poses are used to decouple tracking accuracy and reconstruction quality. Gaussian noise is added to the depth image and camera poses (2mm, 1mm, 0.1$^{\circ}$ standard deviation for depth, translation and orientation, respectively).

\begin{figure*}[t]
\begin{center}
\includegraphics[width=0.9\linewidth]{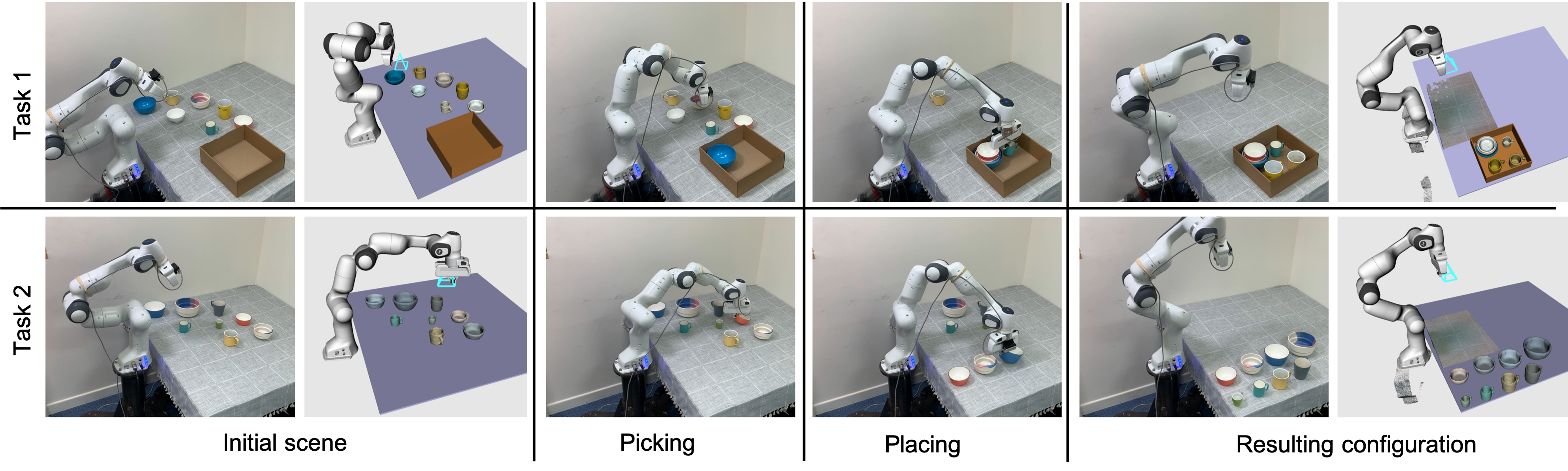}
\end{center}
\caption{Robotic demonstration of packing (task 1) and sorting (task 2) of objects.}
\label{fig:robdemo}
    \vspace{2mm} \hrule
\end{figure*}

\begin{figure}[t]
\begin{center}
   \includegraphics[width=0.85\linewidth]{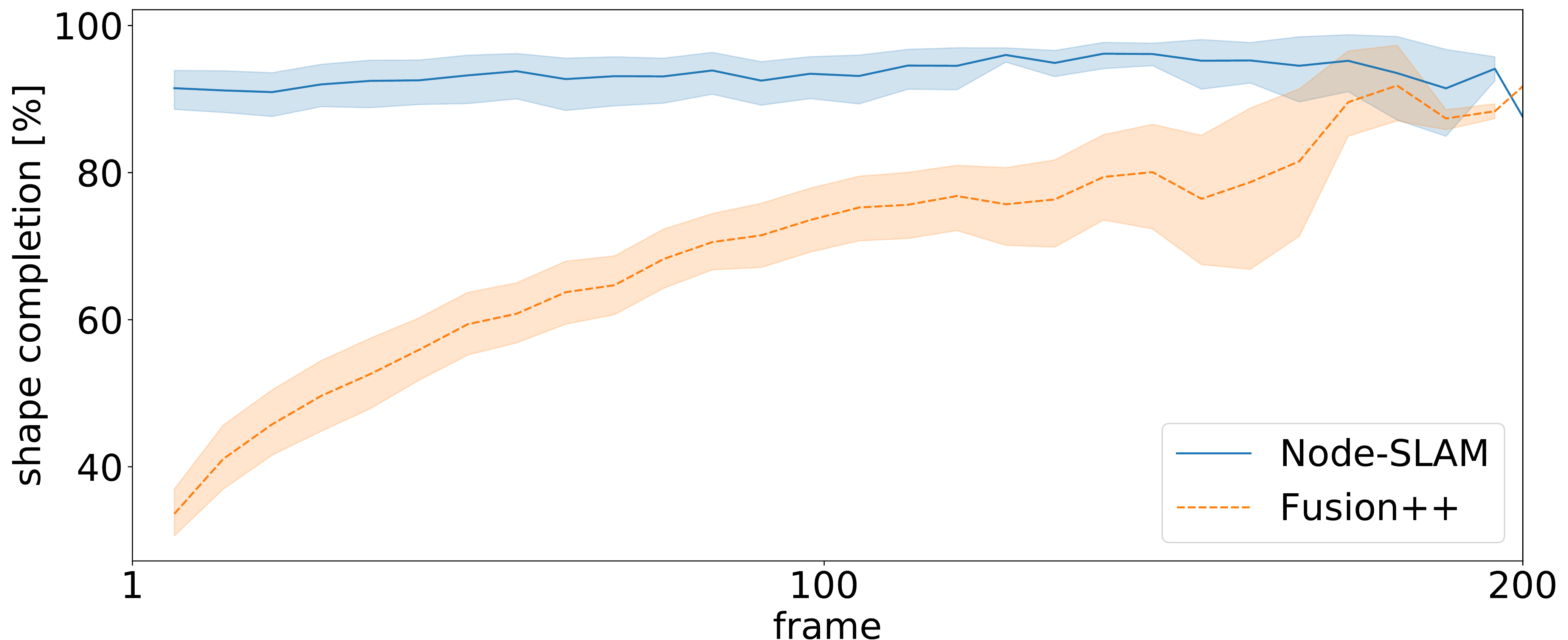}
   \hspace{2mm}
   \includegraphics[width=0.43\linewidth]{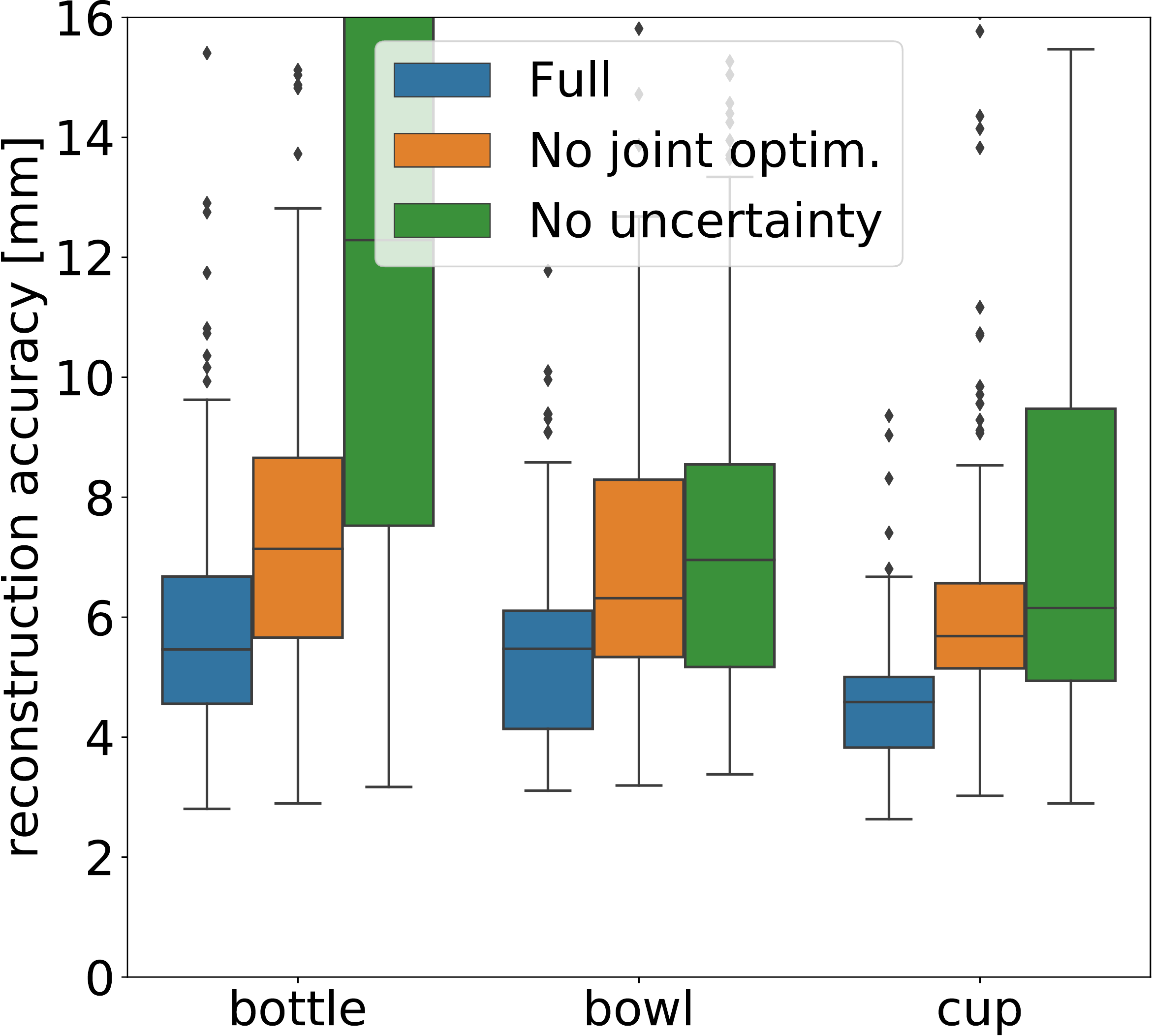}
   \hspace{2mm}
   \includegraphics[width=0.38\linewidth]{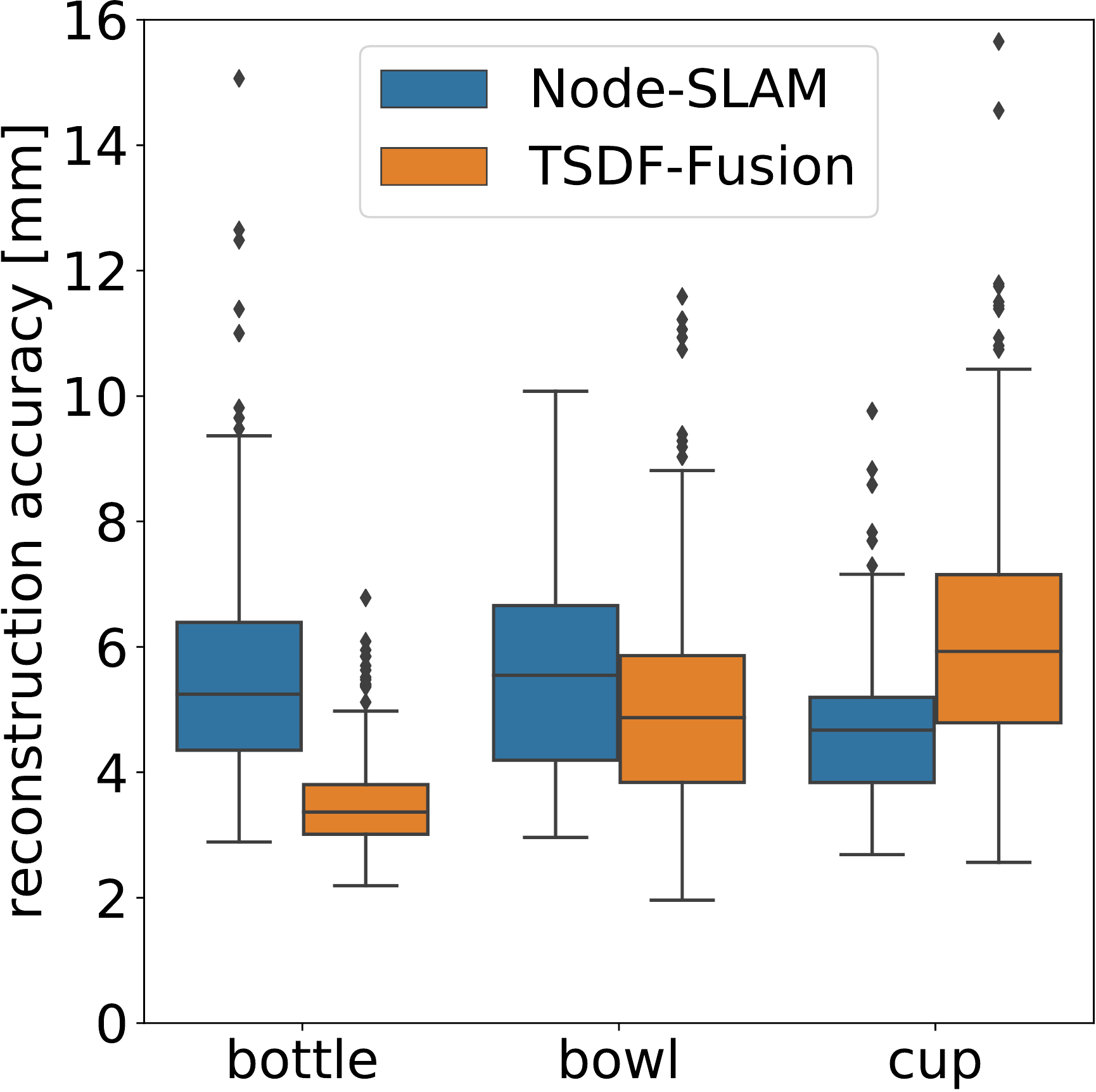}
\end{center}
   \caption {\textbf{Top:} Graph with mean object surface completion (95\% confidence) comparison between NodeSLAM and TSDF fusion, with respect to the number of times an object is updated. \textbf{Bottom left:} Box plots of median surface reconstruction accuracy from our ablation study on 5 scenes with 10 objects in each.  \textbf{Bottom right:} The same metric but comparing our system with Fusion++.}
\label{fig:evals}
    \vspace{2mm} \hrule
\end{figure}

We evaluate shape completion and accuracy; results are accumulated for each class from the 5 simulated sequences.  Figure \ref{fig:evals} shows how mean shape completion evolves with respect to frame number. This graph demonstrates the advantage of class-based priors for object shape reconstruction. With our method we see a jump to almost full completion, while TSDF fusion slowly completes the object with each new fused depth map. Fast shape completion without the need for exhaustive 360 degree scanning is important in robotic applications and in augmented reality, as shown in Figure \ref{fig:ar}.
Figure \ref{fig:evals} displays the median shape accuracy of Node-SLAM compared with TSDF fusion. We observe comparable surface reconstruction quality of close to 5mm.

\subsubsection{Ablation Study}

We evaluate shape reconstruction accuracy and tracking \textit{absolute pose error} on 3 different versions of our system. We compare  our full SLAM system (with camera tracking) with a version without sliding window joint optimisation, and a version without uncertainty rendering.
Figure \ref{fig:evals} shows the importance of these features for shape reconstruction quality, with decreases in performance from 2 up to 7 mm. Table \ref{tab:ape} shows mean \textit{absolute pose error} for each version of our system for all 5 trajectories. These results prove that the precise shape reconstructions from objects provide enough information for accurate camera tracking with mean errors between 1 and 2 cm. It also shows how tracking without joint optimisation or uncertainty leads to significantly lower accuracy on most trajectories.

\begin{table}[t]
 \begin{center}
  \caption{Ablation study for tracking accuracy on 5 scenes, highlighting the importance of a joint optimisation with uncertainty.}
  \label{tab:ape}
    \scriptsize
     \begin{tabular}{ c | c | c | c | c | c }
      \toprule
      \textbf{Absolute Pose} \\ \textbf{Error [cm]} & Scene 1  & Scene 2 & Scene 3 & Scene 4 & Scene 5  \\
      \midrule 
      NodeSLAM &  1.73 & 1 & 0.81 & 1.24 & 1.15  \\
      \midrule 
      NodeSLAM \\ no joint optim. & 8.6 & 10.17 & 0.7 & 2.14 & 1.25  \\
      \midrule 
      NodeSLAM \\ no uncertainty &  4.37  & 3.41 & 0.88 & 3.05 & 6.99  \\
      \bottomrule
    \end{tabular}
  \end{center}
\end{table}

\subsection{Robot Manipulation Application}
We have developed a manipulation application which uses our object reconstruction system. We demonstrate two tasks: object packing and object sorting; see Figure \ref{fig:robdemo} and the attached video. A rapid pre-defined motion is first used to gather a small number of RGB-D views which our system uses to estimate the pose and shape of the objects laid out randomly on a table. Heuristics are used for grasp point selection and a placing motion based on the class and pose of the object and the shape of the reconstructed mesh. All the reconstructed objects are then sorted based on height and radius. For the packing task  all the scanned objects are placed in a tight box, with bowls stacked in decreasing size order and all mugs placed inside the box with centers and orientations aligned. In the sorting task all objects are placed in a line in ascending size. In this robot application only, robot kinematics are used for camera tracking.

\section{Conclusions}

Our generative multi-class object models allow for principled and robust full shape inference. We have shown practical use in a jointly optimisable object SLAM system as well as in two robotic manipulation demonstrations and an augmented reality demo.  We believe that this proves that decomposing a scene into full object entities is a powerful idea for robust mapping and smart interaction. Not all object classes will be well represented by the single code object VAE we used in this paper, and in near future work we plan to investigate alternative coding schemes such as methods which can decompose a complicated object into parts, and methods which can be trained by self-supervision.

\section*{Acknowledgements}
Research presented in this paper has been supported by Dyson Technology Ltd. We thank Michael Bloesch, Shuaifeng Zhi, and Joseph Ortiz for fruitful discussions.

{\small
\bibliographystyle{ieee}
\bibliography{robotvision}

\begin{thebibliography}{10}\itemsep=-1pt

\bibitem{Shapenet:ARXIV2015}
A.~X. Chang, T.~Funkhouser, L.~Guibas, P.~Hanrahan, Q.~Huang, Z.~Li,
  S.~Savarese, M.~Savva, S.~Song, H.~Su, J.~Xiao, L.~Yi, and F.~Yu.
\newblock {ShapeNet}: An information-rich 3d model repository.
\newblock {\em arXiv preprint arXiv:1512.03012}, 2015.

\bibitem{Choy:etal:ECCV2016}
C.~Choy, D.~Xu, J.~Gwak, K.~Chen, and S.~Savarese.
\newblock 3d-r2n2: A unified approach for single and multi-view 3d object
  reconstruction.
\newblock In {\em {Proceedings of the European Conference on Computer Vision
  ({ECCV})}}, 2016.

\bibitem{Amaury:etal:CVPR2013}
A.~Dame, V.~A. Prisacariu, C.~Y. Ren, and I.~Reid.
\newblock Dense reconstruction using 3d object shape priors.
\newblock In {\em {Proceedings of the {IEEE} Conference on Computer Vision and
  Pattern Recognition ({CVPR})}}, 2013.

\bibitem{Davison:etal:PAMI2007}
A.~J. Davison, N.~D. Molton, I.~Reid, and O.~Stasse.
\newblock {{MonoSLAM}: Real-Time Single Camera {SLAM}}.
\newblock {\em {{IEEE} Transactions on Pattern Analysis and Machine
  Intelligence ({PAMI})}}, 29(6):1052--1067, 2007.

\bibitem{Engel:etal:PAMI2017}
J.~Engel, V.~Koltun, and D.~Cremers.
\newblock Direct sparse odometry.
\newblock {\em {{IEEE} Transactions on Pattern Analysis and Machine
  Intelligence ({PAMI})}}, 2017.

\bibitem{Engelmann:etal:WACV2017}
F.~Engelmann, J.~St{\"u}ckler, and B.~Leibe.
\newblock Samp: shape and motion priors for 4d vehicle reconstruction.
\newblock In {\em {Proceedings of the {IEEE} Workshop on Applications of
  Computer Vision ({WACV})}}, 2017.

\bibitem{Gkioxari:etal:ICCV2019}
G.~Gkioxari, J.~Malik, and J.~Johnson.
\newblock Mesh r-cnn.
\newblock 2019.

\bibitem{He:etal:ICCV2017}
K.~He, G.~Gkioxari, P.~Doll{\'a}r, and R.~Girshick.
\newblock Mask r-cnn.
\newblock In {\em {Proceedings of the International Conference on Computer
  Vision ({ICCV})}}, 2017.

\bibitem{Hu:etal:ARXIV2019}
L.~Hu, W.~Xu, K.~Huang, and L.~Kneip.
\newblock Deep-slam++: Object-level rgbd slam based on class-specific deep
  shape priors.
\newblock {\em arXiv preprint arXiv:1907.09691}, 2019.

\bibitem{Izadi:etal:UIST2011}
S.~Izadi, D.~Kim, O.~Hilliges, D.~Molyneaux, R.~A. Newcombe, P.~Kohli,
  J.~Shotton, S.~Hodges, D.~Freeman, A.~J. Davison, and A.~Fitzgibbon.
\newblock {{KinectFusion}: Real-Time {3D} Reconstruction and Interaction Using
  a Moving Depth Camera}.
\newblock In {\em {Proceedings of {ACM} Symposium on User Interface Software
  and Technolog ({UIST})}}, 2011.

\bibitem{Jiang:etal:CVPR2020}
Y.~Jiang, D.~Ji, Z.~Han, and M.~Zwicker.
\newblock Sdfdiff: Differentiable rendering of signed distance fields for 3d
  shape optimization.
\newblock In {\em {Proceedings of the {IEEE} Conference on Computer Vision and
  Pattern Recognition ({CVPR})}}, 2020.

\bibitem{Kajiya:etal:1984}
J.~T. Kajiya and B.~P. Von~Herzen.
\newblock Ray tracing volume densities, 1984.

\bibitem{Kingma:Welling:ICLR2014}
D.~P. Kingma and M.~Welling.
\newblock {Auto-Encoding Variational Bayes}.
\newblock In {\em {Proceedings of the International Conference on Learning
  Representations ({ICLR})}}, 2014.

\bibitem{Klein:Murray:ISMAR2007}
G.~Klein and D.~W. Murray.
\newblock {Parallel Tracking and Mapping for Small {AR} Workspaces}.
\newblock In {\em {Proceedings of the International Symposium on Mixed and
  Augmented Reality ({ISMAR})}}, 2007.

\bibitem{Kundu:etal:CVPR2018}
A.~Kundu, Y.~Li, and J.~M. Rehg.
\newblock 3d-rcnn: Instance-level 3d object reconstruction via
  render-and-compare.
\newblock In {\em {Proceedings of the {IEEE} Conference on Computer Vision and
  Pattern Recognition ({CVPR})}}, 2018.

\bibitem{Li:etal:BMVC2019}
K.~Li, R.~Garg, M.~Cai, and I.~Reid.
\newblock Single-view object shape reconstruction using deep shape prior and
  silhouette.
\newblock 2019.

\bibitem{Li:etal:CVPR2020}
K.~Li, M.~R{\"u}nz, M.~Tang, L.~Ma, C.~Kong, T.~Schmidt, I.~Reid, L.~Agapito,
  J.~Straub, S.~Lovegrove, et~al.
\newblock Frodo: From detections to 3d objects.
\newblock In {\em {Proceedings of the {IEEE} Conference on Computer Vision and
  Pattern Recognition ({CVPR})}}, 2020.

\bibitem{Liu:etal:CVPR2020}
S.~Liu, Y.~Zhang, S.~Peng, B.~Shi, M.~Pollefeys, and Z.~Cui.
\newblock Dist: Rendering deep implicit signed distance function with
  differentiable sphere tracing.
\newblock In {\em {Proceedings of the {IEEE} Conference on Computer Vision and
  Pattern Recognition ({CVPR})}}, 2020.

\bibitem{McCormac:etal:3DV2018}
J.~McCormac, R.~Clark, M.~Bloesch, A.~J. Davison, and S.~Leutenegger.
\newblock {Fusion\texttt{++}}:volumetric object-level slam.
\newblock In {\em {Proceedings of the International Conference on 3D Vision
  ({3DV})}}, 2018.

\bibitem{McCormac:etal:ICRA2017}
J.~McCormac, A.~Handa, A.~J. Davison, and S.~Leutenegger.
\newblock {SemanticFusion}: Dense {3D} semantic mapping with convolutional
  neural networks.
\newblock In {\em {Proceedings of the {IEEE} International Conference on
  Robotics and Automation ({ICRA})}}, 2017.

\bibitem{Mescheder:etal:CVPR2019}
L.~Mescheder, M.~Oechsle, M.~Niemeyer, S.~Nowozin, and A.~Geiger.
\newblock Occupancy networks: Learning 3d reconstruction in function space.
\newblock In {\em {Proceedings of the {IEEE} Conference on Computer Vision and
  Pattern Recognition ({CVPR})}}, 2019.

\bibitem{Newcombe:etal:ISMAR2011}
R.~A. Newcombe, S.~Izadi, O.~Hilliges, D.~Molyneaux, D.~Kim, A.~J. Davison,
  P.~Kohli, J.~Shotton, S.~Hodges, and A.~Fitzgibbon.
\newblock {{KinectFusion}: Real-Time Dense Surface Mapping and Tracking}.
\newblock In {\em {Proceedings of the International Symposium on Mixed and
  Augmented Reality ({ISMAR})}}, 2011.

\bibitem{Niemeyer:etal:CVPR2020}
M.~Niemeyer, L.~Mescheder, M.~Oechsle, and A.~Geiger.
\newblock Differentiable volumetric rendering: Learning implicit 3d
  representations without 3d supervision.
\newblock In {\em {Proceedings of the {IEEE} Conference on Computer Vision and
  Pattern Recognition ({CVPR})}}, 2020.

\bibitem{Park:etal:CVPR2019}
J.~J. Park, P.~Florence, J.~Straub, R.~Newcombe, and S.~Lovegrove.
\newblock Deepsdf: Learning continuous signed distance functions for shape
  representation.
\newblock 2019.

\bibitem{Paschalidou:etal:CVPR2019}
D.~Paschalidou, A.~O. Ulusoy, and A.~Geiger.
\newblock Superquadrics revisited: Learning 3d shape parsing beyond cuboids.
\newblock In {\em {Proceedings of the {IEEE} Conference on Computer Vision and
  Pattern Recognition ({CVPR})}}, 2019.

\bibitem{Russakovsky:etal:ILSVRC15}
O.~Russakovsky, J.~Deng, H.~Su, J.~Krause, S.~Satheesh, S.~Ma, Z.~Huang,
  A.~Karpathy, A.~Khosla, M.~Bernstein, A.~C. Berg, and L.~Fei-Fei.
\newblock {ImageNet Large Scale Visual Recognition Challenge}.
\newblock {\em {International Journal of Computer Vision ({IJCV})}},
  115(3):211--252, 2015.

\bibitem{Salas-Moreno:etal:CVPR2013}
R.~F. Salas-Moreno, R.~A. Newcombe, H.~Strasdat, P.~H.~J. Kelly, and A.~J.
  Davison.
\newblock {{SLAM++}: Simultaneous Localisation and Mapping at the Level of
  Objects}.
\newblock In {\em {Proceedings of the {IEEE} Conference on Computer Vision and
  Pattern Recognition ({CVPR})}}, 2013.

\bibitem{Simonyan:Zisserman:ICLR2015}
K.~Simonyan and A.~Zisserman.
\newblock {Very Deep Convolutional Networks for Large-Scale Image Recognition}.
\newblock In {\em {Proceedings of the International Conference on Learning
  Representations ({ICLR})}}, 2015.

\bibitem{Sohn:etal:NIPS2015}
K.~Sohn, H.~Lee, and X.~Yan.
\newblock Learning structured output representation using deep conditional
  generative models.
\newblock In {\em {Neural Information Processing Systems ({NIPS})}}, 2015.

\bibitem{Sun:etal:CVPR2018}
X.~Sun, J.~Wu, X.~Zhang, Z.~Zhang, C.~Zhang, T.~Xue, J.~B. Tenenbaum, and W.~T.
  Freeman.
\newblock Pix3d: Dataset and methods for single-image 3d shape modeling.
\newblock In {\em {Proceedings of the {IEEE} Conference on Computer Vision and
  Pattern Recognition ({CVPR})}}, 2018.

\bibitem{Sunderhauf:etal:IROS2017}
N.~S\"{u}nderhauf, T.~T. Pham, Y.~Latif, M.~Milford, and I.~Reid.
\newblock Meaningful maps with object-oriented semantic mapping.
\newblock In {\em {Proceedings of the {IEEE/RSJ} Conference on Intelligent
  Robots and Systems ({IROS})}}, 2017.

\bibitem{Tan:etal:CVPR2018}
Q.~Tan, L.~Gao, Y.-K. Lai, and S.~Xia.
\newblock Variational autoencoders for deforming 3d mesh models.
\newblock In {\em {Proceedings of the {IEEE} Conference on Computer Vision and
  Pattern Recognition ({CVPR})}}, 2018.

\bibitem{Tulsiani:etal:CVPR2017}
S.~Tulsiani, T.~Zhou, A.~A. Efros, and J.~Malik.
\newblock Multi-view supervision for single-view reconstruction via
  differentiable ray consistency.
\newblock In {\em {Proceedings of the {IEEE} Conference on Computer Vision and
  Pattern Recognition ({CVPR})}}, 2017.

\bibitem{Wang:etal:ECCV2018}
N.~Wang, Y.~Zhang, Z.~Li, Y.~Fu, W.~Liu, and Y.-G. Jiang.
\newblock Pixel2{M}esh: {G}enerating 3{D} {M}esh {M}odels from {S}ingle {RBG}
  {I}mages.
\newblock In {\em {Proceedings of the European Conference on Computer Vision
  ({ECCV})}}, pages 52--67, 2018.

\bibitem{Wang:etal:ARXIV2019}
R.~Wang, N.~Yang, J.~Stueckler, and D.~Cremers.
\newblock Directshape: Photometric alignment of shape priors for visual vehicle
  pose and shape estimation.
\newblock {\em arXiv preprint arXiv:1904.10097}, 2019.

\bibitem{Whelan:etal:RSS2015}
T.~Whelan, S.~Leutenegger, R.~F. Salas-Moreno, B.~Glocker, and A.~J. Davison.
\newblock {ElasticFusion}: Dense {SLAM} without a pose graph.
\newblock In {\em {Proceedings of Robotics: Science and Systems ({RSS})}},
  2015.

\bibitem{Wu:etal:NIPS2017}
J.~Wu, Y.~Wang, T.~Xue, X.~Sun, B.~Freeman, and J.~Tenenbaum.
\newblock Marrnet: 3d shape reconstruction via 2.5 d sketches.
\newblock In {\em {Neural Information Processing Systems ({NIPS})}}, 2017.

\bibitem{Jiajun:etal:NIPS2016}
J.~Wu, C.~Zhang, T.~Xue, W.~Freeman, and J.~Tenenbaum.
\newblock Learning a probabilistic latent space of object shapes via 3d
  generative-adversarial modeling.
\newblock In {\em {Neural Information Processing Systems ({NIPS})}}, 2016.

\bibitem{Wu:etal:CVPR2015}
Z.~Wu, S.~Song, A.~Khosla, F.~Yu, L.~Zhang, X.~Tang, and J.~Xiao.
\newblock {3D ShapeNets: A Deep Representation for Volumetric Shapes}.
\newblock In {\em {Proceedings of the {IEEE} Conference on Computer Vision and
  Pattern Recognition ({CVPR})}}, 2015.

\bibitem{Xiang:Fox:ARXIV2017}
Y.~Xiang and D.~Fox.
\newblock {DA-RNN}: Semantic mapping with data associated recurrent neural
  networks.
\newblock {\em arXiv preprint arXiv:1703.03098}, 2017.

\bibitem{Yi-Zhou:etal:ARXIV2018}
Q.-Y. Zhou, J.~Park, and V.~Koltun.
\newblock {Open3D}: {A} modern library for {3D} data processing.
\newblock {\em arXiv preprint arXiv:1801.09847}, 2018.

\bibitem{Zhu:etal:WACV2018}
R.~Zhu, C.~Wang, C.-H. Lin, Z.~Wang, and S.~Lucey.
\newblock Object-centric photometric bundle adjustment with deep shape prior.
\newblock In {\em {Proceedings of the {IEEE} Workshop on Applications of
  Computer Vision ({WACV})}}, 2018.

\end{thebibliography}
}

\end{document}